%% file: main.tex
\documentclass[sigconf]{acmart}
\renewcommand\footnotetextcopyrightpermission[1]{}
\usepackage{algorithmic}
\usepackage{algorithm}
\usepackage{amsmath}
\usepackage{multirow}
\usepackage{xcolor}         
\usepackage{color}
\usepackage{colortbl}
\begin{document}

\title{ID-Aligner: Enhancing Identity-Preserving Text-to-Image Generation with Reward Feedback Learning}


\author{Weifeng Chen$^*$}
\email{chenwf35@mail2.sysu.edu.cn}
\affiliation{%
  \institution{Sun Yat-sen University}
  \city{}
  \country{}
}

\author{Jiachang Zhang}
\authornote{Both authors contributed equally. Work done during internship at ByteDance}
\email{zhangjch58@mail2.sysu.edu.cn}
\affiliation{%
  \institution{Sun Yat-sen University}
  \city{}
  \country{}
}

\author{ Jie Wu$^\dagger$}
\email{wujie10558@gmail.com}
\affiliation{%
  \institution{ByteDance Inc.}
  \city{}
  \country{}
}

\author{Hefeng Wu}
\authornote{Corresponding author. }
\email{wuhefeng@gmail.com}
\affiliation{%
  \institution{Sun Yat-sen University}
  \city{}
  \country{}
}

\author{Xuefeng Xiao}
\email{xiaoxuefeng.ailab@bytedance.com}
\affiliation{%
  \institution{ByteDance Inc.}
  \city{}
  \country{}
}

\author{Liang Lin}
\email{linliang@ieee.org}
\affiliation{%
  \institution{Sun Yat-sen University}
  \city{}
  \country{}
}

\renewcommand{\shortauthors}{Chen and Zhang, et al.}

\begin{abstract}
  The rapid development of diffusion models has triggered diverse applications. Identity-preserving text-to-image generation (ID-T2I) particularly has received significant attention due to its wide range of application scenarios like AI portrait and advertising. While existing ID-T2I methods have demonstrated impressive results, several key challenges remain: (1) It is hard to maintain the identity characteristics of reference portraits accurately, (2) The generated images lack aesthetic appeal especially while enforcing identity retention, and (3) There is a limitation that cannot be compatible with LoRA-based and Adapter-based methods simultaneously. To address these issues, we present \textbf{ID-Aligner}, a general feedback learning framework to enhance ID-T2I performance. To resolve identity features lost, we introduce identity consistency reward fine-tuning to utilize the feedback from face detection and recognition models to improve generated identity preservation. Furthermore, we propose identity aesthetic reward fine-tuning leveraging rewards from human-annotated preference data and automatically constructed feedback on character structure generation to provide aesthetic tuning signals. Thanks to its universal feedback fine-tuning framework, our method can be readily applied to both LoRA and Adapter models, achieving consistent performance gains. Extensive experiments on SD1.5 and SDXL diffusion models validate the effectiveness of our approach. \textbf{Project Page: \url{https://idaligner.github.io/}}
\end{abstract}

\keywords{Text-to-Image Generation, Diffusion Model, Feedback Learning, Identity-Preserving Generation}

\begin{teaserfigure}
  \includegraphics[width=0.99\textwidth]{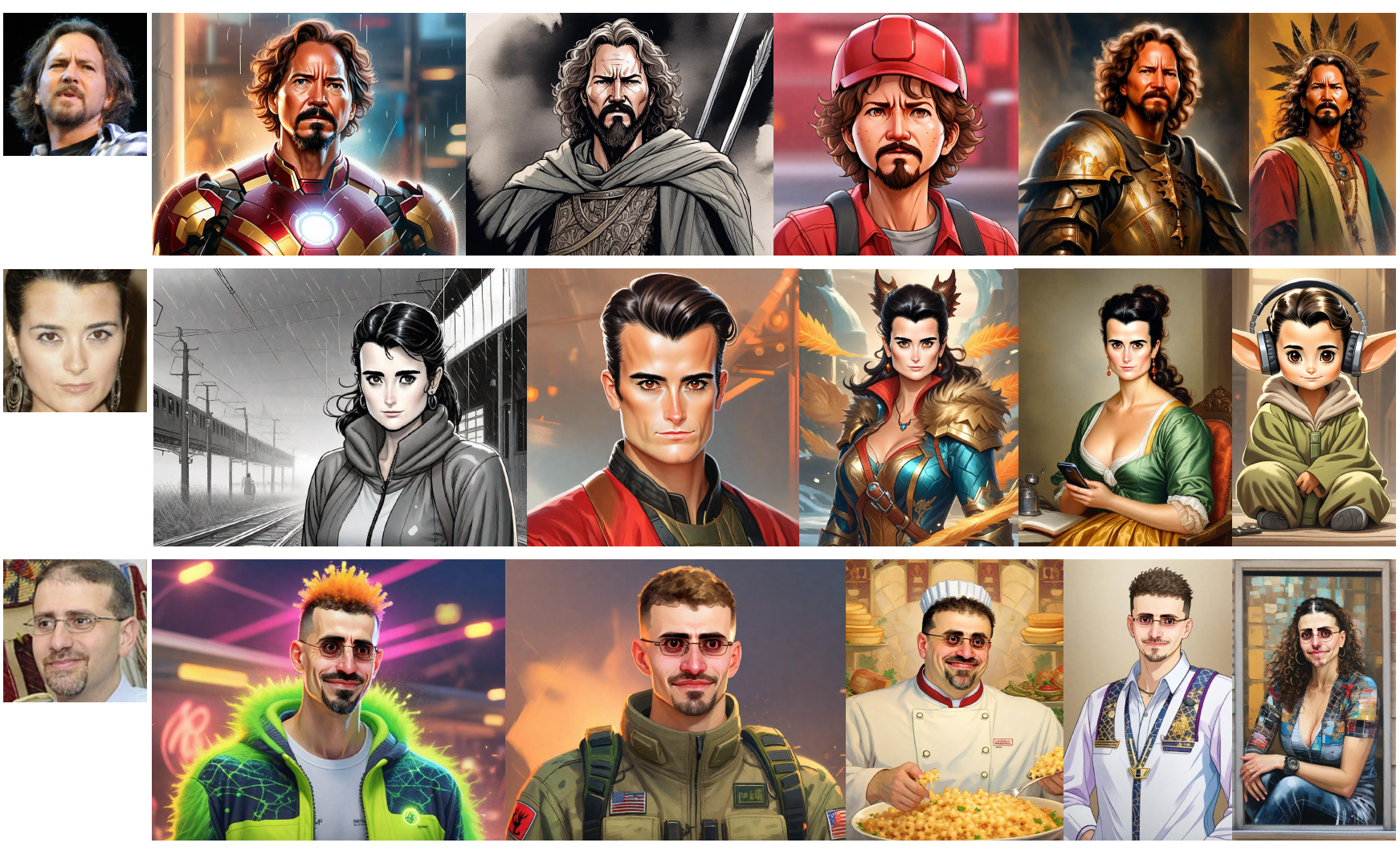}
  \caption{We present \textit{ID-Aligner}, a general framework to boost the performance of identity-preserving text-to-image generation from the feedback learning perspective. We introduce an identity consistency reward and an identity aesthetic reward to enhance the identity preservation and the visual appeal of the generated characters. Our approach can apply to both the LoRA-based and Adapter-based methods, and exhibit superior performance compared with existing methods. }
    \Description{Enjoying the baseball game from the third-base
  seats. Ichiro Suzuki preparing to bat. }
  \label{fig:teaser}
\end{teaserfigure}

\maketitle

\input{sec/1_intro}

\input{sec/2_related}

\input{sec/3_method}

\input{sec/4_exp}
\input{sec/5_conclusion}

\bibliographystyle{ACM-Reference-Format}
\bibliography{sample-base}

\clearpage
\appendix


\section{More Results}

\subsection{Generalization}
\begin{figure*}
    \centering
    \includegraphics[width=0.99\linewidth]{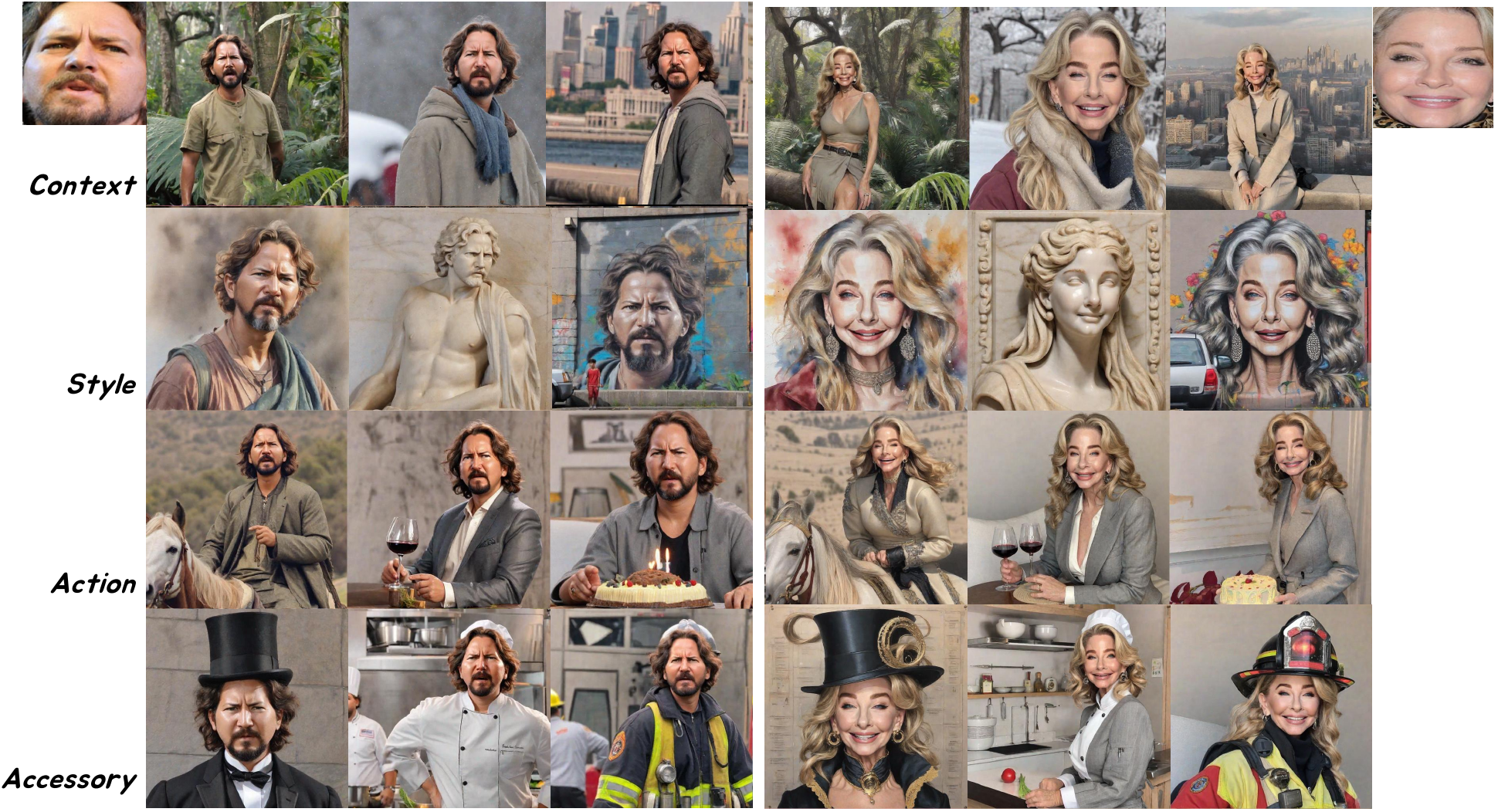}
    \caption{Visualizations of generation results obtained by using prompt in Tab.\ref{tab:prompt_table}. It shows prompts of different types including "context", "style", "action" and "accesory".}
    \label{fig:prompt_vis}
\end{figure*}

Our evaluation strategy focuses on the crucial aspect of generalization, ensuring that our model's capabilities extend beyond specific scenarios. To achieve this, we leverage a diverse array of prompts categorized into four distinct groups. These prompts provide a comprehensive assessment of the model's proficiency in preserving identity across varied contexts. The specifics of our evaluation prompts are meticulously documented in Table \ref{tab:prompt_table}, offering transparency and reproducibility in our methodology. In Figure \ref{fig:prompt_vis}, we visually showcase the outputs generated by our model across these varied prompt categories. From depicting individuals in different environmental contexts ("context" prompts) to showcasing diverse attire choices ("accessory" prompts), engaging in various activities ("action" prompts), and adopting different stylistic elements ("style" prompts), our model consistently demonstrates its versatility and adaptability. This comprehensive exploration underscores the robustness of our approach in handling a wide spectrum of identity-preserving generation tasks. By effectively navigating through diverse prompts, our model exemplifies its ability to capture the essence of individual identity across a multitude of scenarios, thereby showcasing its potential for real-world applications and further advancing the field of generative models.

\subsection{Identity Mixing}
\begin{figure*}[h]
    \centering
    \includegraphics[width=1\linewidth]{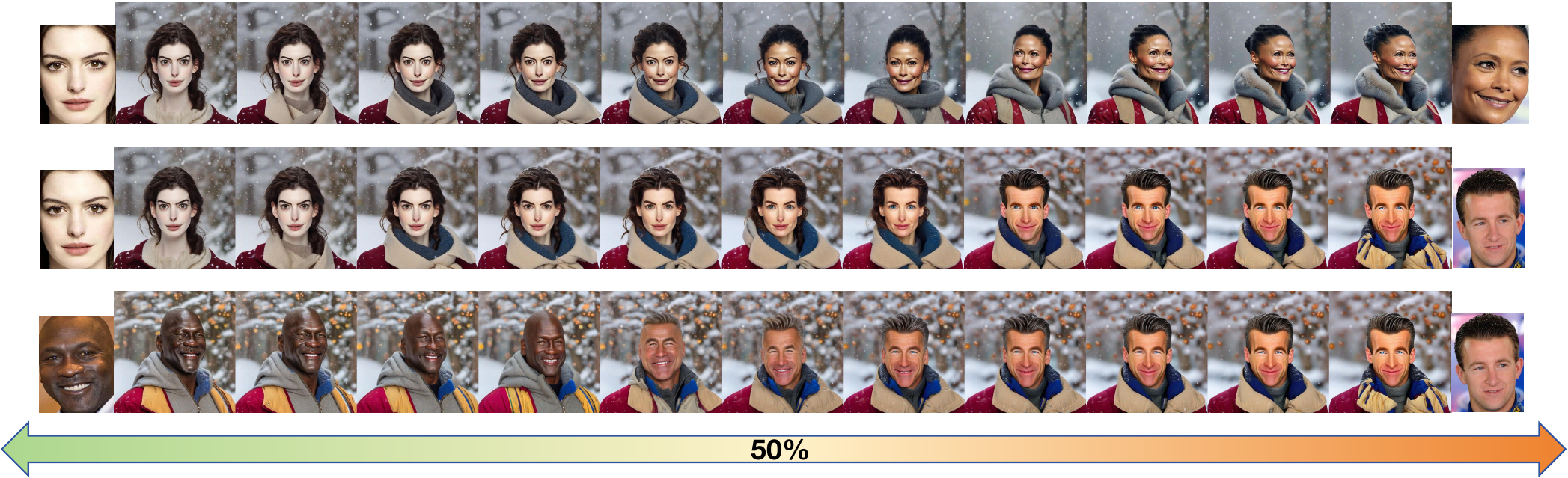}
    \caption{Visualizations of generation results obtained by incorporating our optimized IP-Adapter by mixing two people as image condition. The prompt is "a person in snowy day, closeup". The mixing rate range from 0 to 1 with 0.1 as interval.}
    \label{fig:idmixing}
\end{figure*}

We further explore the application of our method in identity mixing. Specifically, given 2 different identity reference images, we aim to generate the synthesized image that contains the characteristics of both identities. This poses a higher demand for the model’s identity-preserving ability. We show some results of our method in identity mixing in Fig. \ref{fig:idmixing}. We provide a white lady mixing with another black lady, a white lady mixing with another yellow man, black man mixing with another yellow man to show the generalization of our model.
\subsection{More Comparison with Other Methods}
\begin{figure*}[h]
    \centering
    \includegraphics[width=0.8\linewidth]{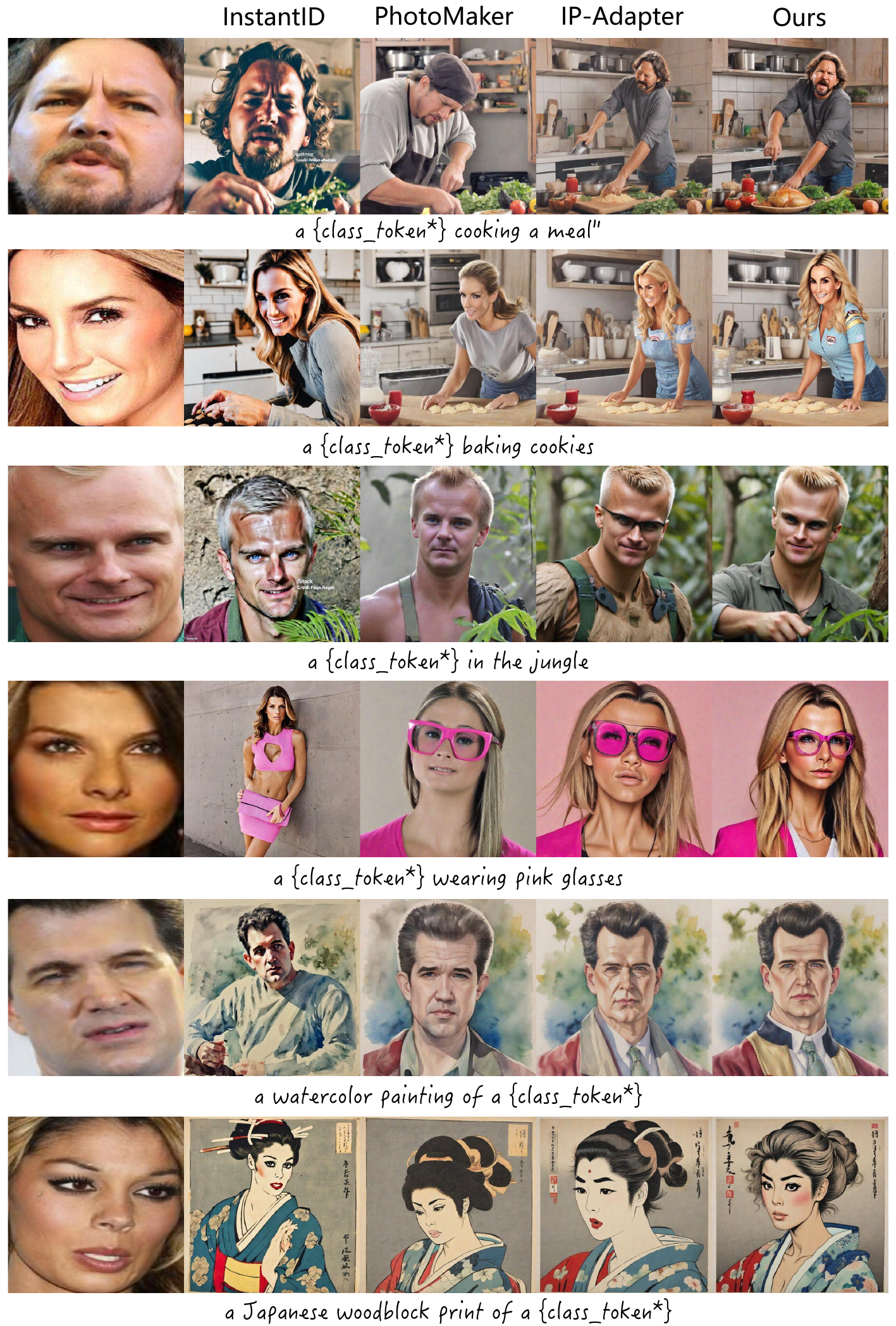}
    \caption{Visualizations of generation results obtained by incorporating our optimized IP-Adapter into more open-source text-to-image base models in Civitai.}
    \label{fig:more_visualization}
\end{figure*}

We present additional visualization results of both other methods and our own in Figure \ref{fig:more_visualization}. It is evident that our method outperforms in terms of identity-preserving generation, exhibiting superior performance in both identity consistency and visual appeal. It is worth noting that the InstantID method \cite{instantid} achieves better identity preservation by incorporating an additional IdentityNet. This approach is similar to ControlNet \cite{controlnet} and imposes a strict condition on the original reference. However, this constraint often leads to the generation of rigid images that closely follow the exact pose of the reference image. As a result, some unreasonable results are produced, as exemplified by the case of "a \{class\_token\} cooking a meal". In contrast, our method consistently generates more visually pleasing results, which have a better trade-off in terms of text consistency and identity consistency. 


\section{Limitations}

\noindent(1) Our algorithm is designed to enhance existing models through reward feedback learning. However, if the existing model is already robust, the improvement may be marginal.

\noindent(2) Enhancing face similarity may occasionally compromise prompt consistency, as the emphasis on facial features might lead to undesirable outcomes. This issue can be mitigated by reducing the intensity of identity control.

\noindent(3) Biases inherent in T2I models and their training data can impact results. Certain identities may yield more favorable outcomes, while others may produce dissimilar results.

\section{Broader Impact}

The impact of our research extends across multiple dimensions. Academically, our method serves as a foundational framework for integrating diffusion models with other expert models, such as face recognition models. This integration contributes significantly to the advancement of generative models.
Practically, our technique holds immense transformative potential across a wide spectrum of industries, including entertainment, portraiture, advertising, and beyond. By providing a means to generate high-quality human images with fidelity, our approach offers unprecedented opportunities for creativity and innovation.
Nevertheless, it is essential to recognize and address the ethical considerations inherent in the widespread adoption of such technology. The capacity to produce lifelike human images raises legitimate concerns regarding privacy, potential misuse, and the dissemination of false information. Thus, we underscore the critical importance of developing and adhering to stringent ethical guidelines to ensure the responsible and ethical utilization of this groundbreaking technology.


\begin{table*}[!h]
\centering
\label{tab:model}
\caption{The selected open-sourced text-to-image models in the Civitai community.}
\begin{tabular}{cl}
\hline
\textbf{Model} & \textbf{URL} \\ \hline
Realistic Vision v60-b1 & \href{https://civitai.com/models/4201/realistic-vision-v60-b1}{https://civitai.com/models/4201/realistic-vision-v60-b1} \\ 
DreamShaper & \href{https://civitai.com/models/4384/dreamshaper}{https://civitai.com/models/4384/dreamshaper} \\ 
GhostMix & \href{https://civitai.com/models/36520/ghostmix}{https://civitai.com/models/36520/ghostmix} \\ 
ToonYou & \href{https://civitai.com/models/30240/toonyou}{https://civitai.com/models/30240/toonyou} \\ 
Disney Pixar Cartoon TypeB & \href{https://civitai.com/models/75650/disney-pixar-cartoon-typeb}{https://civitai.com/models/75650/disney-pixar-cartoon-typeb} \\ 
Disney Stylev1 & \href{https://civitai.com/models/114413/disney-stylev1}{https://civitai.com/models/114413/disney-stylev1} \\ 
\hline
\end{tabular}
\end{table*}

\begin{table*}[!h]
  \centering
  \begin{minipage}[t]{0.4\textwidth}
    \centering
    \footnotesize
   \begin{tabular}{l|c}
    \toprule
    Category & Prompt \\
    \midrule
    \multirow{10}{*}{Accessory} & "a \{class\_token\} wearing a red hat" \\
        & "a \{class\_token\} wearing a Santa hat"  \\
        &  "a \{class\_token\} wearing a rainbow scarf"  \\
       &   "a \{class\_token\} wearing a black top hat and a monocle"  \\
        &  "a \{class\_token\} in a chef outfit"  \\
        &  "a \{class\_token\} in a firefighter outfit"  \\
        &  "a \{class\_token\} in a police outfit"  \\
        &  "a \{class\_token\} wearing pink glasses"  \\
        &  "a \{class\_token\} wearing a yellow shirt"  \\
        &  "a \{class\_token\} in a purple wizard outfit" \\
    \hline
    \multirow{10}{*}{Style} &     "a painting of a \{class\_token\} in the style of Banksy" \\
         &  "a painting of a \{class\_token\} in the style of Vincent Van Gogh" \\
         &  "a colorful graffiti painting of a \{class\_token\}" \\
         &  "a watercolor painting of a \{class\_token\}" \\
         &  "a Greek marble sculpture of a \{class\_token\}" \\
         &  "a street art mural of a \{class\_token\}" \\
         &  "a black and white photograph of a \{class\_token\}" \\
         &  "a pointillism painting of a \{class\_token\}" \\
         &  "a Japanese woodblock print of a \{class\_token\}" \\
         &  "a street art stencil of a \{class\_token\}" \\
    \bottomrule
    \end{tabular}
  \end{minipage}
  \hfill
  \begin{minipage}[t]{0.4\textwidth}
    \centering
    \footnotesize
     \begin{tabular}{l|c}
    \toprule
    Category & Prompt \\
    \midrule
    \multirow{10}{*}{Cotenxt} & "a \{class\_token\} in the jungle" \\
     &  "a \{class\_token\} in the snow" \\
     &  "a \{class\_token\} on the beach" \\
     &  "a \{class\_token\} on a cobblestone street" \\
     &  "a \{class\_token\} on top of pink fabric" \\
     &  "a \{class\_token\} on top of a wooden floor" \\
     &  "a \{class\_token\} with a city in the background" \\
     &  "a \{class\_token\} with a mountain in the background" \\
     &  "a \{class\_token\} with a blue house in the background" \\
     &  "a \{class\_token\} on top of a purple rug in a forest" \\
    \hline
    \multirow{10}{*}{Action} &     "a \{class\_token\} riding a horse" \\
     & "a \{class\_token\} holding a glass of wine" \\
     & "a \{class\_token\} holding a piece of cake" \\
     & "a \{class\_token\} giving a lecture" \\
     & "a \{class\_token\} reading a book" \\
     & "a \{class\_token\} gardening in the backyard" \\
     & "a \{class\_token\} cooking a meal" \\
     & "a \{class\_token\} working out at the gym" \\
     & "a \{class\_token\} walking the dog" \\
     & "a \{class\_token\} baking cookies" \\
    \bottomrule
    \end{tabular}
 \end{minipage}
    \caption{The prompt used in the evaluation procedure of the adapter model. Following \cite{fastcomposer}, we exploit diverse prompts of four distinct categories to evaluate the performance of our method in identity-preserving generation comprehensively.}
    \label{tab:prompt_table}
    
\end{table*}

\end{document}

%% file: sec/1_intro.tex
\section{Introduction}
\label{sec:intro}
In recent years,  the field of image synthesis has experienced a remarkable revolution with the emergence of diffusion models. These powerful generative diffusion models, exemplified by significant milestones such as DALLE-2 \cite{dalle-2} and Imagen \cite{imagen}, have completely reshaped the landscape of text-to-image (T2I) generation. Moreover, the development of these models has also given rise to many related application tasks, such as image editing \cite{sdedit}, controllable image generation \cite{controlnet,unicontrol}, and so on. Among these, the identity-preserving text-to-image generation received widespread attention due to its broad application scenarios like E-commerce advertising, AI portraits, image animation, and virtual try-it-on. It aims to generate new images about the identity of a specific person in the reference images with the guidance of the textual prompt. There are numerous advanced research works on this task. Early works resort to the low-rank (LoRA) \cite{lora} to adapt the pre-trained text-to-image diffusion model to the given a few reference portrait images and achieve recontextualization of the particular identity. Recently, IP-Adapter \cite{ipadapter} achieved impressive personalized portrait generation by inserting an adapter model into the attention module of the diffusion model and fine-tuning using a high-quality large-scale dataset of facial images. However, despite these achievements, these methods still fall short in several aspects: (i) They cannot achieve accurate identity preservation. Existing methods typically employ a mean squared error (MSE) loss during training, which is unable to explicitly learn image generation that faithfully captures the characteristics of the reference portrait as shown in Fig.\ref{fig:id_aligner_ablation}. (ii) The generated image tends to lack appeal, especially when enforcing identity consistency. For example, the state-of-the-art method InstantID \cite{instantid} introduces an extra IdentityNet to retain the information of reference portrait. While high fidelity, such a strict constraint is also prone to generating rigid images or characters with distorted/unnatural limbs and poses as depicted in Fig.\ref{fig:comparison}. (iii) Existing methods either rely on LoRA \cite{lora} or Adapter \cite{ipadapter} to achieve ID-T2I generation and lack a general method that is compatible with these two paradigms. \par
In this work, drawing inspiration from the recent advancements in feedback learning within the diffusion model \cite{imagereward,ddpo,hps}, we present \textbf{ID-Aligner}, a framework to boost the identity image generation performance with specially designed reward models via feedback learning. Specifically, we introduce an identity consistency reward tuning to boost identity preservation. It employs the face detection model along with the face recognition model as the reward model to measure identity consistency and provide specialized feedback on identity consistency, which enables superior identity consistency during the recontextualization of the portrait in the reference images. In addition, to enhance the aesthetic quality of the identity generation, we further introduce an identity aesthetic reward tuning, which exploits a tailored reward model trained with human-annotated preference feedback data and automatically constructs character structure feedback data to steer the model toward the aesthetic appealing generation. Our method is very flexible and can be applied to not only the adapter-based model but also the LoRA-based model and achieve a consistent performance boost in both identity preservation and aesthetic quality. We also observe the significant acceleration effect with the LoRA-based model, facilitating its wide application. Extensive experiments demonstrate the superiority of our method upon the existing method, such as IP-Adapter \cite{ipadapter}, PhotoMaker \cite{photomaker}, and InstantID \cite{instantid}.
Our contributions are summarized as follows:

\begin{itemize}
\item  We present ID-Aligner, a general feedback learning framework to improve the performance of identity-preserving text-to-image generation in both identity consistency and aesthetic quality. To the best of our knowledge, this is the first work to address this task through feedback learning.

\item We introduce a universal method that can be applied to both the LoRA-based model and the Adapter-based model. Theoretically, our approach can boost all the existing training-based identity-preserving text-to-image generation methods.

\item Extensive experiments have been conducted with various existing methods such as IP-Adapter, PhotoMaker, and InstanceID, validating the effectiveness of our method in improving identity consistency and aesthetic quality.
\end{itemize}

%% file: sec/2_related.tex
\section{Related Works}
\noindent\textbf{Text-to-Image Diffusion Models.~}
Recently,  diffusion models \cite{ddim, ddpm} have showcased remarkable capabilities in the realm of text-to-image (T2I) generation. Groundbreaking works like Imagen \cite{imagen}, GLIDE \cite{glide}, and DALL-E2 \cite{dalle-2} have emerged, revolutionizing text-driven image synthesis and reshaping the landscape of the T2I task. Notably, the LDM \cite{ldm} model, also known as the stable diffusion model, has transformed the diffusion process from pixel space to latent space, significantly improving the efficiency of training and inference. Building upon this innovation, the Stable Diffusion XL (SDXL) \cite{sdxl} model has further enhanced training strategies and achieved unprecedented image generation quality through parameter scaling. The development of these models has also triggered various applications, including image editing \cite{sdedit, brooks2022instructpix2pix, hertz2022prompt, tumanyan2022plug}, controllable image generation \cite{controlnet, t2iadpater, composer}, etc.

\begin{figure*}[!t]
    \centering
    \includegraphics[width=0.95\linewidth]{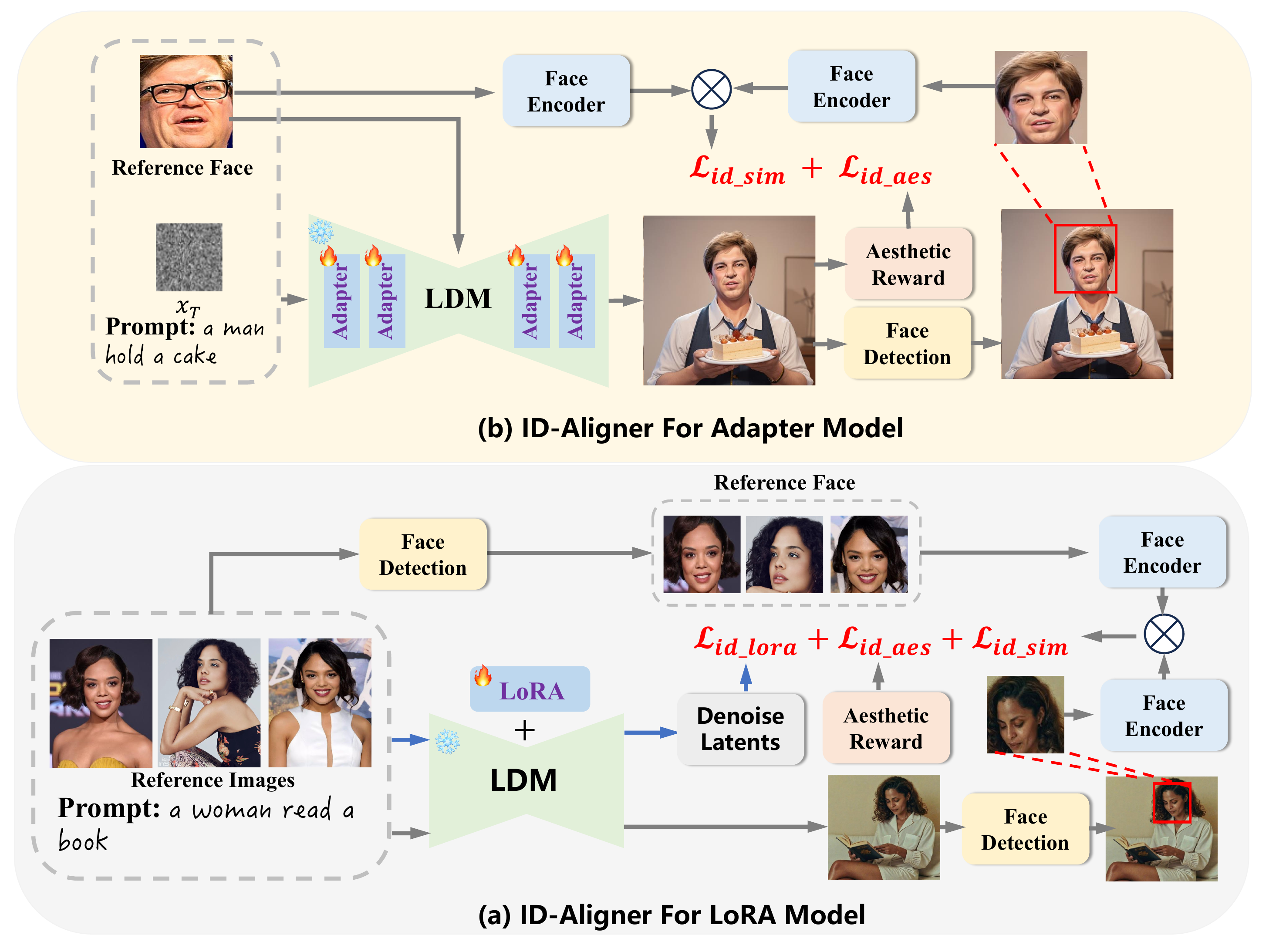}
    \vspace{-6pt}
    \caption{The overview of the proposed \textbf{\textit{ID-Aligner}}. Our method exploits face detection and face encoder to achieve identity preservation via feedback learning. We further incorporated the aesthetic reward model to improve the visual appeal of the generation results. Our method is a general framework that can be applied to both LoRA and Adapter methods.}
    \label{fig:pipeline}
\end{figure*}

\noindent\textbf{Identity-Preserving Image Generation.~}
ID-preserving image Generation \cite{instantbooth, facestudio, photomaker, photoverse, instantid, ipadapter} has emerged as a significant application of text-to-image generation, capturing widespread attention due to its diverse range of application scenarios. The primary objective of this application is to generate novel images about a particular identity of one or several reference images guided by textual prompts. Unlike the conventional text-generated image task, it is crucial not only to ensure the performance of prompt comprehension and generation fidelity but also to maintain consistency of the ID information between the newly generated images and the reference images. Few-shot methods \cite{lora, dreambooth, instantbooth, elite, subjectdiffusion} attempted to to finetune the diffusion model given several reference images to learn the ID features. However, this approach requires specific fine-tuning for each character, which limits its flexibility. PhotoMaker \cite{photomaker} achieves ID preservation by fine-tuning part of the Transformer layers in the image encoder and merging the class and image embeddings. IP-Adapter-FaceID \cite{ipadapter} uses face ID embedding in the face recognition model instead of CLIP \cite{clip} image embedding to maintain ID consistency. Similarly, InstantID \cite{instantid} uses a FaceEncoder to extract semantic Face Embedding and inject the ID information via Decoupled Cross-Attention. In addition, an IdentityNet is designed to introduce additional spatial control. In contrast to these approaches, our method relies on feedback learning, eliminating the need for intricate network structures. It offers exceptional versatility and effectiveness, seamlessly adapting to various existing methods while significantly enhancing ID Preservation.

\noindent\textbf{Human Feedback for Diffusion Models.~}
Inspired by the success of reinforcement learning with human feedback (RLHF) in the field of Large Language Models (LLMs)  \cite{chatgpt, rlhf, instructgpt}, researchers \cite{ddpo, hps, hpsv2} have tried to introduce feedback-based learning into the field of text-to-image generation. Among these, DDPO \cite{ddpo} employs reinforcement learning to align the diffusion models with the supervision provided by the additional reward models. Different from DDPO, HPS \cite{hps,hpsv2} exploits the reward model trained on the collected preference data to filter the preferred data and then achieve feedback learning via a supervised fine-tuning manner. Recently, ImageReward \cite{imagereward} proposes a ReFL framework to achieve preference fine-tuning, which performs reward scoring on denoised images within a predetermined diffusion model denoising step range through a pre-trained reward model, backpropagates and updates the diffusion model parameters. Recently, UniFL \cite{unifl} proposes a unified framework to enhance diffusion models via feedback learning. Inspire by these, in this paper, we propose a reward feedback learning algorithm that focuses on optimizing ID-T2I models.

%% file: sec/3_method.tex
\section{Method}
We introduce ID-Aligner, a pioneering approach that utilizes the feedback learning method to enhance the performance of identity (ID) preserving generation. The outline of our method can be seen in Fig. \ref{fig:pipeline}. We resolve the ID-preserving generation via a reward feedback learning paradigm to enhance the consistency with a reference face image and aesthetic of generated images. 


\subsection{Text-to-Image Diffusion Model}

Text-to-image diffusion models leverage diffusion modeling to generate high-quality images based on textual prompts via the diffusion model, which generates desired data samples from Gaussian noise through a gradual denoising process. 
During pre-training, a sampled image $x$ is first processed by a pre-trained VAE \cite{vae, taming} encoder to derive its latent representation $z$. 
Subsequently, random noise is injected into the latent representation through a forward diffusion process, following a predefined schedule $\{\beta_t\}^T$. 
This process can be formulated as $z_t = \sqrt{\overline{{\alpha}}_t} z + \sqrt{1-\overline{{\alpha}}_t}\epsilon$, where $\epsilon \in \mathcal N(0, 1)$ is the random noise with identical dimension to $z$, $\overline{{\alpha}}_t = \prod_{s=1}^t\alpha_s$ and $\alpha_t = 1 - \beta_t$. To achieve the denoising process, a UNet $\epsilon_{\theta}$ is trained to predict the added noise in the forward diffusion process, conditioned on the noised latent and the text prompt $c$. Formally, the optimization objective of the UNet is:
\begin{equation}\label{eq:add_noise}\begin{aligned}
\mathcal L(\theta) = \mathbb E_{z,\epsilon,c,t}[||\epsilon - \epsilon_{\theta}(\sqrt{\overline{{\alpha}}_t} z + \sqrt{1 - \overline{{\alpha}}_t}\epsilon, c,t)||_2^2 ] .
\end{aligned}
\end{equation}

\subsection{Identity Reward}

\noindent\textbf{Identity Consistency Reward:} Given the reference image $x^{\text{ref}}_0$ and the generated image $x_0'$. Our objective is to assess the ID similarity of the particular portrait. To achieve this, we first employ the face detection model \texttt{FaceDet} to locate the faces in both images. Based on the outputs of the face detection model, we crop the corresponding face regions and feed them into the encoder of a face recognition model $\texttt{FaceEnc}$. This allows us to obtain the encoded face embeddings for the reference face $\mathcal E_{\text{ref}}$ and the generated face $\mathcal E_{\text{gen}}$, i.e.,
\begin{align}
\mathcal E_{\text{ref}}=\texttt{FaceEnc}(\texttt{FaceDet}(x^{\text{ref}}_0)), \\
\mathcal E_{\text{gen}}=\texttt{FaceEnc}(\texttt{FaceDet}(x_0')).
\label{eq:}
\end{align}

Subsequently, we calculate the cosine similarity between these two face embeddings, which serves as the measure of ID retention during the generation process. We then consider this similarity as the reward signal for the feedback tuning process as follows:
\begin{equation}
\Re_{id\_sim}(x_0', x^{\text{ref}}_0) = \texttt{cose\_sim}(\mathcal E_{\text{gen}}, \mathcal E_{\text{ref}}).
\label{eq:idsim}
\end{equation}


\begin{figure}[ht]
    \centering
    \includegraphics[width=0.99\linewidth]{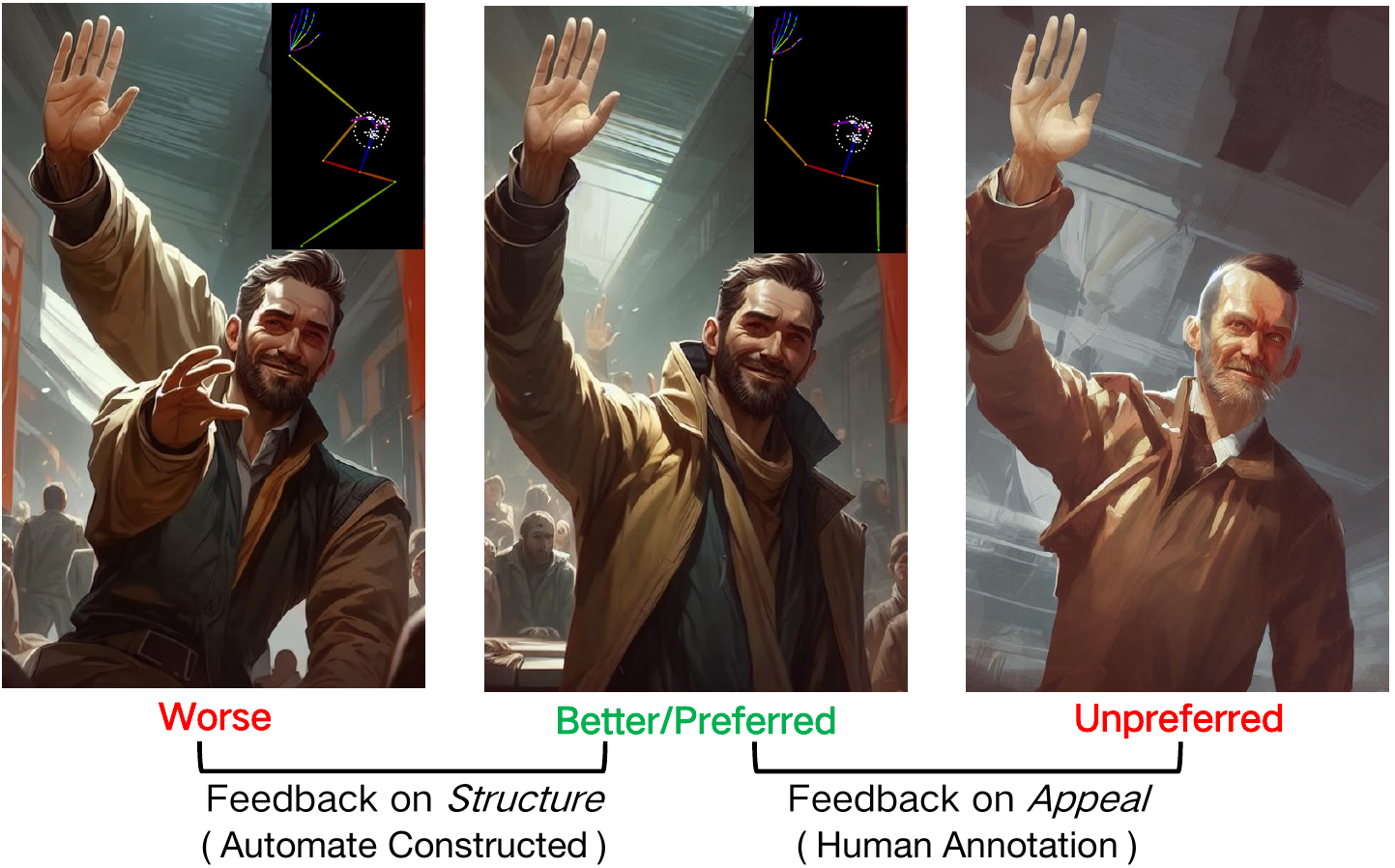}
    \caption{The illustration of the aesthetic feedback data construction. We take an ``AI + Expert'' way to generate the feedback data. \textbf{Left}: The automatic data construction for the feedback data on the character structure generation. We resort to ControlNet \cite{controlnet} to manually generate the structure-distorted negative samples. \textbf{Right:} Human annoatated preference data over images. }
    \vspace{-5pt}
    \label{fig:feeback_data}
\end{figure}

\noindent\textbf{Identity Aesthetic Reward:} In addition to the identity consistency reward, we introduce an identity aesthetic  reward model focusing on appealing and quality. It consists of human preference of appeal and a reasonable structure. 

First, we train a reward model with self-collected human annotation preference dataset that can score the image and reflect human preference over the appeal, as is shown in right of Fig.\ref{fig:feeback_data}. We employ the pretrained model provided by ImageReward \cite{imagereward} and finetune it with the following loss: 
\begin{equation}
    \begin{aligned}
        \mathcal{L}_{\theta}=-E_{(c,x_i,x_j) \sim \mathcal{D} } [ log(\sigma(f_{\theta}(x_i,c) - f_{\theta}(x_j,c))) ].
    \label{eq:reward_training}
    \end{aligned}
\end{equation}
This loss function is based on comparison pairs between images, where each comparison pair contains two images ($x_i$ and $x_j$) and prompt $c$. $f_\theta(x, c)$ represents the reward score given an image $x$ and a prompt $c$. We therefore term $f_\theta$ as $\Re_{appeal}$ for appealing reward.

In addition, we design a structure reward model that can distinguish distorted limbs/body from natural one. To train a model that can access the whether the structure of image is reasonable or not, we collect a set of text-image pairs containing positive and negative samples. Specifically, we use the images from LAION \cite{laion} filtered with human detector. We then use a pose estimation model to generate a pose, which can be treat as undistored human structure. We then randomly twiste the pose and utilize ControlNet \cite{controlnet} to generate distored body as negative samples, as is shown in left of Fig.\ref{fig:feeback_data}. Once the positive and negative pairs are available, similarly, we train the structure reward model with the same loss of Eq. \ref{eq:reward_training} as well and term structure reward model as $\Re_{struct}$. 

Then, the identity aesthetic reward model is defined as
\begin{equation}
\Re_{id\_aes}(x, c) = \Re_{appeal}(x, c) + \Re_{struct}(x, c).
\label{eq:idsim}
\end{equation}



\begin{algorithm}[t]
    \renewcommand{\algorithmicrequire}{\textbf{Input:}}
    \renewcommand{\algorithmicensure}{\textbf{Output:}}
    \caption{\footnotesize ID-Preserving Reward Feedback Learning for \textbf{Adapter model}}
    \begin{algorithmic}[1]
          \STATE \textbf{Dataset:} Identity preservation generation text-image dataset $\mathcal{D} = \{ (\textrm{txt}_1, \textrm{ref\_face}_1), ... (\textrm{txt}_n, \textrm{ref\_face}_n) \}$
        \STATE \textbf{Input:} LDM with pre-trained adapter parameters $w_0$, face detection model $\texttt{FaceDet}$, encoder of a face recognition model  $\texttt{FaceEnc.}$
        \STATE \textbf{Initialization:} The number of noise scheduler time steps $T$, add noise timestep $T_a$, denoising time step $t$.
        \FOR {data point $(\textrm{txt}_i, \textrm{ref\_face}_i) \in \mathcal{D}$}
           
            \STATE $x_T$ $\gets$ RandNoise  // Sample a Guassion noise.
            \STATE $t$ $\gets$ Rand($T_1$, $T_2$) // Pick a random denoise time step $t \in [T_1, T_2]$
            
            \FOR {$j$ = $T$, ..., $t+1$}
                \STATE \textbf{no grad:} $x_{j-1}$ $\gets$ $\textrm{LDM}_{w_i}$$\{x_{j} |(\textrm{txt}_i, \textrm{ref\_face}_i)\}$
            \ENDFOR
            \STATE \textbf{with grad:} $x_{t-1} \gets \textrm{LDM}_{w_i}\{x_{t}|(\textrm{txt}_i, \textrm{ref\_face}_i)\}$
            \STATE $x^{'}_0 \gets x_{t-1}$ // Predict the original latent by noise scheduler
             \STATE $\textrm{img}^{'}_i$ $\gets$ VaeDec($x^{'}_0$) // From latent to image
            \STATE $\textrm{a}^{'}_i$ $\gets$  $\texttt{FaceDet}$($\textrm{img}^{'}_0$) // Detect the face area in the denoised image
             \STATE $\textrm{emb}^{'}_i$,  $\textrm{emb}_i$   $\gets$ $\texttt{FaceEnc}$($\textrm{a}^{'}_i$), $\texttt{FaceEnc}$($\textrm{ref\_face}_i$) // Extract the embedding of generated face and reference face
            \STATE  $\mathcal{L}_{id\_reward} $ $\gets$ $\mathcal{L}_{id\_sim}$($\textrm{emb}^{'}_i$,  $\textrm{emb}_i$) + $\mathcal{L}_{id\_aes}$($\textrm{img}^{'}_i$)   // ID reward loss
            \STATE $w_{i+1} \gets w_i$ // Update $\textrm{Adapter}_{w_i}$ 
        \ENDFOR
    \end{algorithmic}  
    \label{adapter_algo}
\end{algorithm}

\subsection{ID-Preserving Feedback Learning}

In the feedback learning phase, we begin with an input prompt $c$, initializing a latent variable $x_T$ at random. The latent variable is then progressively denoised until reaching a randomly selected timestep $t$. At this point, the denoised image $x^{\prime}_0$ is directly predicted from $x_t$. The reward model obtained from the previous phase is applied to this denoised image, generating the expected preference score. This preference score enables the fine-tuning of the diffusion model to align more closely with our ID-Reward that reflects identity consistency and aesthetic preferences:
\begin{equation}
        \mathcal{L}_{id\_sim} = \mathbb E_{c \sim p(c)}\mathbb E_{x^{\prime}_0 \sim p(x^{\prime}_0|c)}[1 - \Re_{id\_sim}(x_0', x_0^{ref})],
\end{equation}
\begin{equation}\label{eq:refl}
\mathcal{L}_{id\_aes} = \mathbb E_{c \sim p(c)}\mathbb E_{x^{\prime}_0 \sim p(x^{\prime}_0|c)}[-\Re_{id\_aes}(x^{\prime}_0,c)].
\end{equation}
%

Finally, we use the weighted sum of these two reward objectives to fine-tune the diffusion for ID-preserving image generation:
\begin{equation}
     \begin{aligned}
         \mathcal L_{id\_reward} = \alpha_1 \mathcal L_{id\_sim} + \alpha_2 \mathcal L_{id\_aes},
     \end{aligned}
     \label{eq:idreward}
\end{equation}
where $\alpha_1$ and $\alpha_2$ are the balancing coefficients.

Our \textbf{ID-Aligner} is a universal method that can be applied to both the LoRA-based model and the Adapter-based model for ID-preserving generation, as described below in detail.

\vspace{3pt}\noindent\textbf{ID-Aligner For Adapter Model.~~}
IP-Adapter is pluggable model for diffusion model, which enable a face image as identity control. We optimize this model with reward feedback learning, as shown in Fig.\ref{fig:pipeline}(a). We follow the same spirits of ReFL~ \cite{imagereward} to utilize a tailor reward model to provide a special feedback signal on identity consistency. Specifically, given a reference image of a particular portrait and a textual control prompt, $(x^{\text{ref}}_0, p)$, we first iteratively denoise a randomly initialized latent without gradient until a random time step $T_d \in [DT1, DT2]$, yielding $x_{T_d}$. Then, a further denoise step is executed with a gradient to obtain $x_{T_d - 1}$ and directly get the predicted denoised image $x_0'$ from $x_{T_d-1}$. Afterward, a reward model is utilized to score on $x_0'$ and steer the model toward to the particular direction according to the reward model guidance. Here, we use weighted sum of similarity reward $L_{id\_sim}$ and aesthetic reward $L_{id\_aes}$ to fetch loss $L_{id\_reward}$ in Eq.~\ref{eq:idreward} to optimize the model. The complete process is summarized in Algorithm~\ref{adapter_algo}.


\vspace{3pt}\noindent\textbf{ID-Aligner For LoRA Model.~~}
LoRA is an efficient way to achieve identity-preserving generation. Given single or several reference images of the particular portrait, it quickly adapts the pre-trained LDM to the specified identity by solely fine-tuning some pluggable extra low-rank parameters matrix of the network. However, few-shot learning in diffusion model to learn to generate a new person is highly depends on the provided dataset, which may require faces from different aspect or environment to avoid over-fitting. In this paper, we propose a more efficient way for ID LoRA training by applying the mentioned ID reward. As is shown in Fig.\ref{fig:pipeline}(b), we train the LoRA with  weighted sum of a denoising loss $L_{id\_lora}$ in Eq.\ref{eq:add_noise} and ID-reward loss $L_{id\_reward}$ in Eq.\ref{eq:idreward}. The $L_{id\_lora}$ enables the model to learn the face structure while the $L_{id\_sim}$ guide the model to learn identity information. The extra $L_{id\_aes}$ is applied for improving the overall aesthetic of images. The complete process is summarized in Algorithm~\ref{lora_algo}, which is slightly difference from the adapter one in terms of loss design.

\begin{algorithm}[t]
    \renewcommand{\algorithmicrequire}{\textbf{Input:}}
    \renewcommand{\algorithmicensure}{\textbf{Output:}}
    \caption{\footnotesize ID-Preserving Reward Feedback Learning for \textbf{LoRA model}}
    \begin{algorithmic}[1]
          \STATE \textbf{Dataset:} Several personalized text-image pairs dataset $\mathcal{D} = \{ (\textrm{txt}_1, \textrm{img}_1), ... (\textrm{txt}_n, \textrm{img}_n) \}$
        \STATE \textbf{Input:} LDM with LoRA parameters $w_0$, face detection model $\texttt{FaceDet}$, encoder of a face recognition model  $\texttt{FaceEnc.}$
        \STATE \textbf{Initialization:} The number of noise scheduler time steps $T$, add noise timestep $T_a$, denoising time step $t$. 
        
        \STATE $\textrm{emb}_{ref}$  $\gets$ $\texttt{Average}$($\texttt{FaceEnc}$($\texttt{FaceDet}$($\textrm{img}_i$))), $i$ $\in$ $\mathcal{D}$ // extract ID embeddings of personalized images. 
        
        \FOR {data point $(\textrm{txt}_i, \textrm{img}_i) \in \mathcal{D}$}
           
            \STATE $x_T$ $\gets$ RandNoise  // Sample a Guassion noise.
            \STATE $x_l$ // Add noise into the latent $x_0$ according to Eq.\ref{eq:add_noise}

            // Denoising
            \STATE \textbf{with grad:} $x_{l-1} \gets \textrm{LDM}_{w_i}\{x_{l}|(\textrm{txt}_i)\}$

            // ID-Reward Loop
            \STATE $t$ $\gets$ Rand($T_1$, $T_2$) // Pick a random denoise time step $t \in [T_1, T_2]$
            
            \FOR {$j$ = $T$, ..., $t+1$}
                \STATE \textbf{no grad:} $x_{j-1}$ $\gets$ $\textrm{LDM}_{w_i}$$\{x_{j} |(\textrm{txt}_i)\}$
            \ENDFOR
            \STATE \textbf{with grad:} $x_{t-1} \gets \textrm{LDM}_{w_i}\{x_{t}|(\textrm{txt}_i)\}$
            \STATE $x^{'}_0 \gets x_{t-1}$ // Predict the original latent by noise scheduler
             \STATE $\textrm{img}^{'}_i$ $\gets$ VaeDec($x^{'}_0$) // From latent to image
            \STATE $\textrm{a}^{'}_i$ $\gets$  $\texttt{FaceDet}$($\textrm{img}^{'}_0$) // Detect the face area in the denoised image
             \STATE $\textrm{emb}^{'}_i$   $\gets$ $\texttt{FaceEnc}$($\textrm{a}^{'}_i$)  // Extract the embedding of generated face
            \STATE  $\mathcal{L}_{id\_reward} $ $\gets$ $\mathcal{L}_{id\_sim}$($\textrm{emb}^{'}_i$,  $\textrm{emb}_{ref}$) + $\mathcal{L}_{id\_aes}$($\textrm{img}^{'}_i$)   + 
            $\mathcal{L}_{mse} $ ( $x_{l-1}$ , $x_0$ ) // ID reward loss + denoising MSE loss
            \STATE $w_{i+1} \gets w_i$ // Update $\textrm{LoRA}_{w_i}$ 
        \ENDFOR
    \end{algorithmic}  
    \label{lora_algo}
\end{algorithm}    

%% file: sec/4_exp.tex
\begin{figure*}
     \centering
     \includegraphics[width=0.98\linewidth]{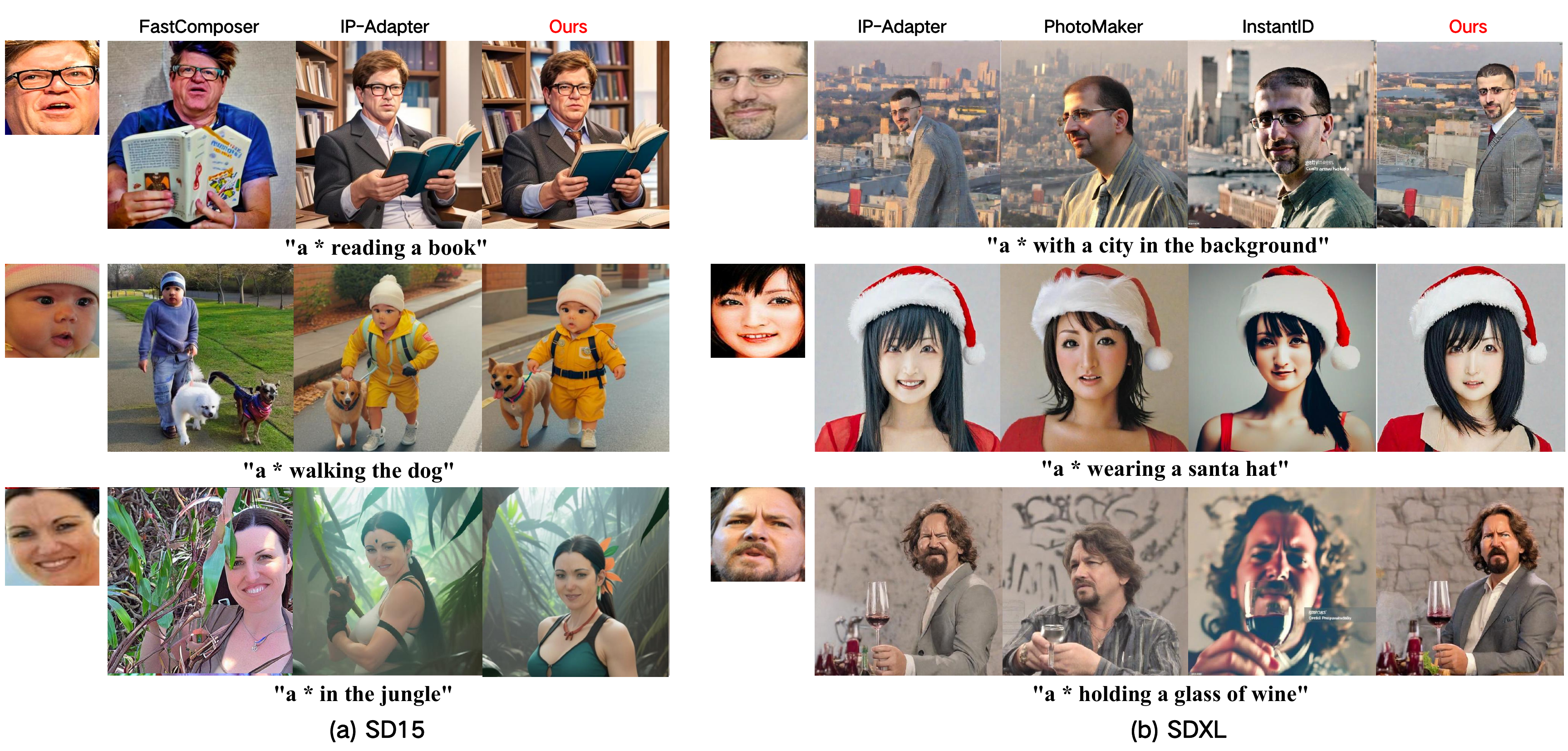}
     \caption{Visual comparison of different Adapter-based identity conditional generation methods based on SD15 and SDXL.}
     \label{fig:comparison}
\end{figure*}

\section{Experiments}\label{exp_sec}

\textbf{Traning Dataset:} We carefully curated a portrait dataset specifically for ID-preserving generation training. Specifically, we employed the MTCNN face detector \cite{mtcnn} to filter the image from the LAION dataset \cite{laion}. This process finally resulted in over 200,000 images that contained faces. These images were used for both LoRA and Adapter training. For adapter-fashion training, we cropped the face from each image and used it as the reference identity. To enhance the model's generalization capability, we further collect a high-quality prompt set from JourneyDB \cite{pan2023journeydb} for identity-conditioned generation. To make sure the prompt is compatible with the ID-preserving generation task, we manually filter the prompt containing the referring word about the identity summarized by chatGPT\cite{chatgpt}, such as 'the girl', 'the man', which finally yielded a final set of prompts describing human.

\noindent\textbf{Training \& Inference Details:} For \textbf{Adapter} model, we take \texttt{stable-diffusion-v1-5} \cite{ldm} and $\texttt{SDXL}$\cite{sdxl} as the base text-to-image generation models, and take the widely recognized IP-Adapter \cite{ipadpter} as the baseline model. We initialize the adapter weights with the pre-trained weight of $\texttt{IP-Adapter-faceid\_plusv2} $ \cite{ipadpter}. During training, only the adapter parameters are updated, ensuring compatibility with other models of the same structure. The model is trained using the 512x512 (1024x1024 for SDXL) resolution image with a batch size of 32. The learning rate is $10^{-6}$, and totally trained for 10,000 iterations. Following the practice of 
 \cite{imagereward}, the guidance scale is set to 1.0 during feedback learning. The $\alpha_1$ is set as 0.2 and $\alpha_2$ is set as 0.001. As for \textbf{LoRA} model, we collect 5 images for each identity. We use bucket adaptive resolution during LoRA training with a batch size of 1. The learning rate is $5 * 10^{-5}$ for LoRA layers of UNet and $1 * 10^{-4}$ for LoRA layers of Text encoder. The LoRA training is based on \texttt{stable-diffusion-v1-5} \cite{ldm} and \texttt{SDXL} \cite{sdxl} and totally trained for 2,000 iterations. For both LoRA and Adapter training, we exploit FaceNet \cite{facenet} as the face detection model and MTCNN \cite{mtcnn} face recognition model for the face embedding extraction. During inference, the DDIM scheduler \cite{ddim} is employed, sampling 20 steps for generation. The guidance scale is set to 7.0, and the fusion strength of the adapter module is fixed at 1.0.

\noindent\textbf{Evaluation settings:} We evaluate the identity-preserving generation ability of our method with the prompt from the validation set of FastComposer\cite{fastcomposer}. These prompts encompass four distinct types, namely action behavior, style words, clothing accessories, and environment. These diverse prompts facilitate a comprehensive evaluation of the model's capacity to retain the identity throughout the generation process. For the LoRA model, we collect 5 images for 6 characters ranging from black, white, and yellow skin.  Separate LoRA is trained for each character, and the performance is evaluated individually. In the case of the adapter model, we carefully gather an image collection of about 20 portraits from various sources including existing face datasets and the internet. This collection represents a rich spectrum of identities, spanning various genders such as men and women, ages encompassing both adults and children, and diverse skin tones including black, white, and yellow. These images serve as conditional reference images during the evaluation process. Following \cite{photomaker}, we report the face similarity score (similarity between the generated face and the reference face), DINO score (similarity between the perceptual representation of the generated image and the reference image), CLIP-I (semantic similarity of generated image and reference images), and CLIP-T (semantic similarity of text prompt and the generated images) to evaluate the performance of our method.

\begin{table*}
\centering
\caption{Quantitative comparison between the state-of-the-art methods. The best results are highlighted in bold, while the second-best results results are underlined.}
\label{tab:quantitative_results}
    \vspace{-3pt}
\begin{tabular}{lcccccc}
\hline
Architecture &Model  & Face Sim.$\uparrow$ & DINO$\uparrow$  & CLIP-I$\uparrow$  & LAION-Aes$\uparrow$ & CLIP-T$\uparrow$ \\
\hline

\multirow{4}{*}{SD1.5} & FastComposer & 0.486 & 0.498 & 0.616 & 5.44 & \textbf{24.0} \\
& IP-Adapter & \underline{0.739} & \underline{0.586} & \underline{0.684} & \underline{5.54} & \underline{22.0} \\
& Ours & \textbf{0.800} & \textbf{0.606} & \textbf{0.727} & \textbf{5.59} & 20.6 \\

\hline
\multirow{5}{*}{SDXL} & IP-Adapter & 0.512 & 0.460 & 0.541 & \underline{5.85} & \textbf{24.5} \\
& InstantID & \textbf{0.783} & 0.445 & \underline{0.606} & 5.58 & 22.8 \\
& PhotoMaker & 0.520 & \underline{0.497} & \textbf{0.641} & 5.54 & 23.6 \\
& Ours & \underline{0.619} & \textbf{0.499} & 0.602 & \textbf{5.88} & \underline{23.7} \\
\hline
\end{tabular}
\end{table*}

\subsection{Experimental results}
\subsubsection{\textbf{Qualitative Comparison}}

We conduct qualitative experiment of ID-Aligner for IP-Adapter and Lora model. 

\noindent\textbf{Adapter Model:} We compare our model's performance with baseline methods and other state-of-the-art adapter-based models. As illustrated in Figure \ref{fig:comparison}, we conduct experiments on both the SD15 and SDXL models. In Figure \ref{fig:comparison}(a), which showcases the results for the SD15-based model, our method demonstrates superior identity preservation and aesthetic quality. For instance, in the second example of "a * walking the dog," both FastComposer and IP-Adapter fail to generate a reasonable image of a baby, resulting in lower image quality. The third example highlights our method's ability to better preserve the identity, aligning closely with the reference face. Regarding the SDXL-based model in \ref{fig:comparison}(b), InstantID exhibits the best capability for identity-preserving generation. However, it has lower flexibility due to the face ControlNet, where the generated images heavily rely on the input control map. For example, in the case of "a * holding a glass of wine," InstantID only generates an avatar photo, while other methods can produce half-body images without the constraint of the face structure control map. We show competitive face similarity with it. Meanwhile, our method have better aesthetic that any other methods, the clarity and aesthetic appeal is better than other, for example, the color of the second case and the concrete structure of the third case.

\begin{figure}[!t]
    \centering
    \includegraphics[width=1\linewidth]{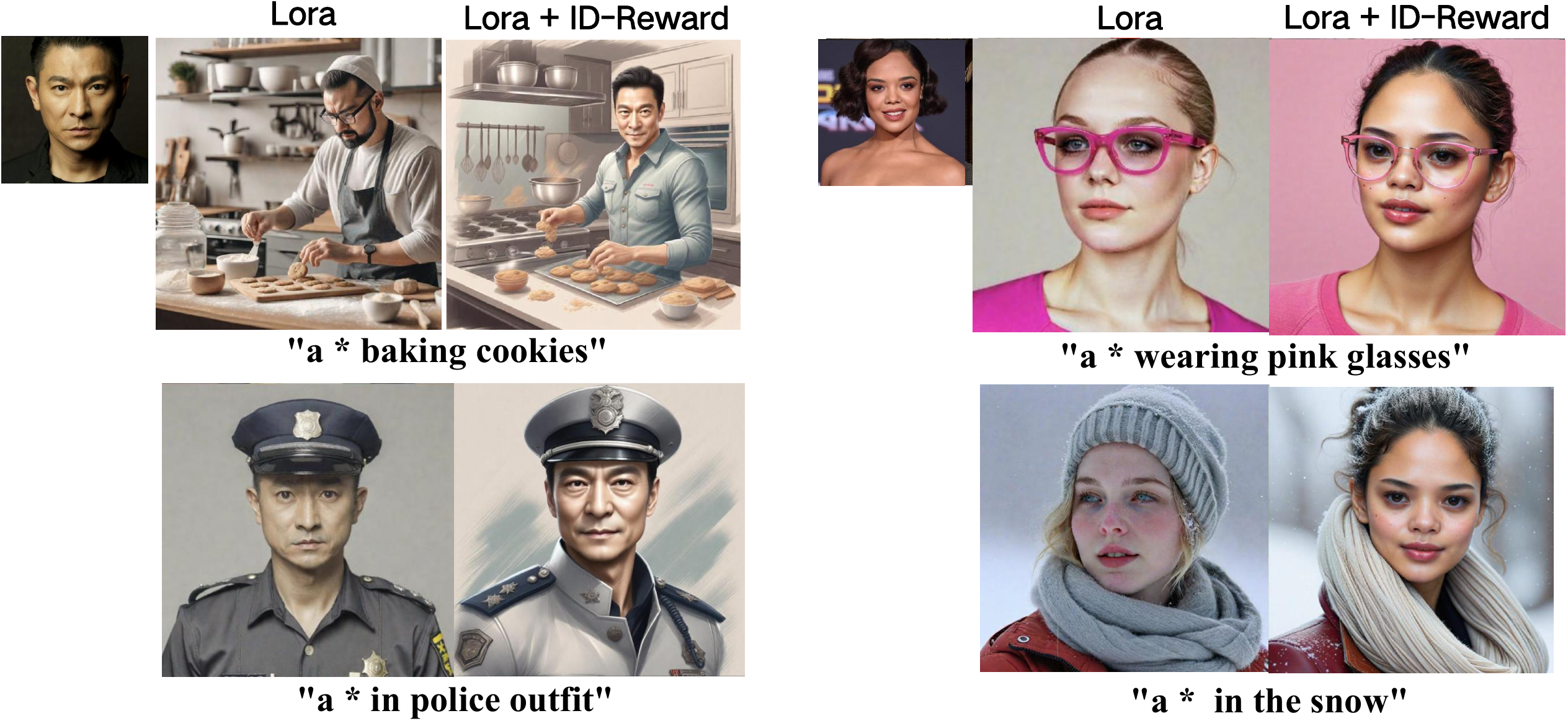}
    \caption{Visual results of LoRA ID-Aligner methods based on SDXL.}
    \label{fig:id_aligner_lora_results}
\end{figure}

\noindent\textbf{LoRA Model:} Fig.\ref{fig:id_aligner_lora_results} showcase the results of incorporating our method into the LoRA model. It is evident that our method significantly boosts identity consistency (the male character case) and visual appeal (the female character case) compared with the naive LoRA method. 

\subsubsection{\textbf{Quantitative Comparison}}
Tab.\ref{tab:quantitative_results} presents a quantitative comparison of our proposed method with several state-of-the-art techniques for identity-preserving image generation, evaluated across various metrics. The methods are categorized based on the underlying architecture, with results reported separately for the SD1.5 and SDXL models. For the SD1.5 model, our method outperforms FastComposer and IP-Adapter in terms of Face Similarity (Face Sim.), DINO, and CLIP-I scores, indicating superior identity preservation consistency. Specifically, our approach achieves a Face Sim. score of 0.800, surpassing IP-Adapter's 0.739 and FastComposer's 0.486, suggesting better face identity preservation. Additionally, our higher DINO (0.606) and CLIP-I (0.727) scores demonstrate improved overall subject consistency. Our method also yields the highest LAION-Aesthetics (LAION-Aes) score of 5.59, indicating enhanced aesthetic quality compared to the baselines. Regarding the SDXL model, InstantID exhibits the highest Face Sim. score of 0.783, outperforming our method (0.619) and the other baselines in terms of face identity preservation. However, our approach achieves competitive performance on the DINO (0.499) and CLIP-I (0.602) metrics, suggesting comparable overall identity consistency. Notably, our method obtains the highest LAION-Aes score of 5.88 among all SDXL-based techniques, demonstrating its ability to generate aesthetically pleasing images while maintaining identity consistency. We also note that there is a slight performance drop in the semantic alignment between the prompt and the generated image after the optimization. This is because the model is forced to focus on identity adaptation, and will inevitably overlook the textual prompt to some extent. This phenomenon is also observed in lots of existing identity preservation generation works \cite{photomaker,dreambooth,instantid}.

\begin{table*}
\centering
\caption{Generalization study of our method on different base T2I models: \textbf{Dreamshaper} (SD1.5) and \textbf{RealVisXL} (SDXL).}
\label{tab:generalization_study}
\begin{tabular}{lcccc}
\hline
Model & Face Sim.$\uparrow$ & DINO$\uparrow$  & CLIP-I$\uparrow$ \\
\hline
IP-Adapter-Dreamshaper & 0.598 & 0.583 & 0.591 \\
\rowcolor{gray!10}IP-Adapter-Dreamshaper + ID-Reward & \textbf{0.662 \textcolor[RGB]{96,177,87}{(+10.7\%)}} & \textbf{0.588 \textcolor[RGB]{96,177,87}{(+0.8\%)}}  & \textbf{0.616 \textcolor[RGB]{96,177,87}{(+4.2\%)}} \\
IP-Adapter-RealVisXL & 0.519 & 0.488 & 0.575 \\
\rowcolor{gray!10}IP-Adapter-RealVisXL + ID-Reward& \textbf{0.635 \textcolor[RGB]{96,177,87}{(+22.3\%)}} & \textbf{0.509 \textcolor[RGB]{96,177,87}{(+4.3\%)}} & \textbf{0.623 \textcolor[RGB]{96,177,87}{(+8.3\%)}} \\
\hline
\end{tabular}
\end{table*}

\subsubsection{\textbf{Ablation Study}}

\begin{figure}
    \centering
    \includegraphics[width=1\linewidth]{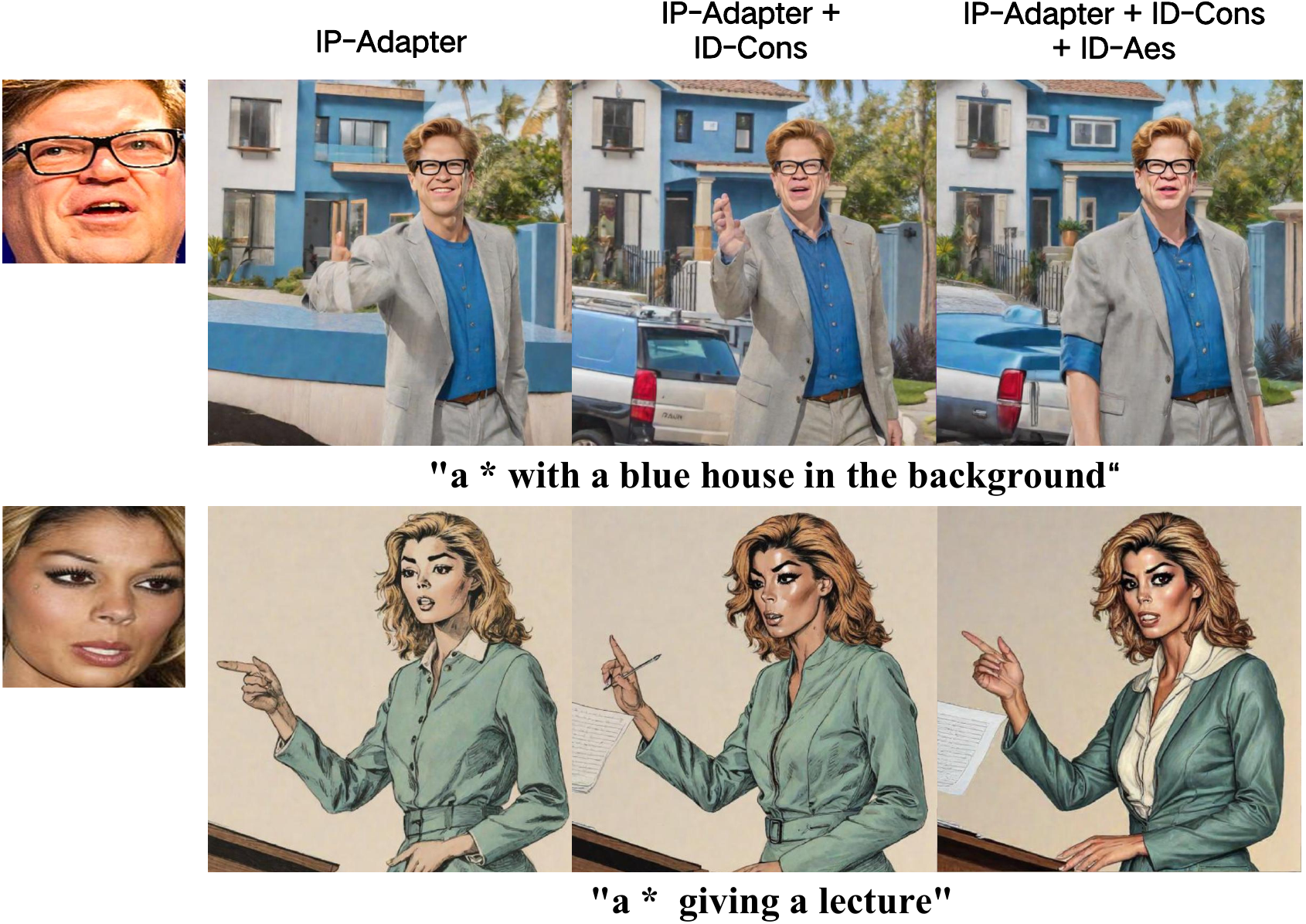}
    \caption{The effectiveness ablation of the proposed identity consistency reward (ID-Cons) and the identity aesthetic reward (ID-Aes).}
    \label{fig:id_aligner_ablation}
\end{figure}

We conduct ablation study to analyze the effectiveness of each component in our method.

\noindent\textbf{Identity Reward:} We conduct an ablation study to evaluate the impact of the proposed identity consistency reward and aesthetic reward. As illustrated in Fig.\ref{fig:id_aligner_ablation}, applying the identity consistency reward boosts the identity similarity significantly. For example, both the two cases generated by the baseline model encounters severe identity loss with a notably different appearance from the reference portrait.  However, after optimizing the model using the identity consistency reward, the generated character exhibits a more similar outlook. This improvement can be attributed to the specialized identity reward provided by the face detection and face embedding extraction model, which guides the model to preserve the desired identity features as much as possible. 
Furthermore, the incorporation of the identity aesthetic reward further enhances the visual appeal of the generated image, particularly in improving the structural aspects of the characters. 
For example, in the first row, despite the preservation of identity features achieved through the identity consistency reward, the generated hands of the character still exhibit distortion. However, this issue is effectively resolved by the identity aesthetic reward, which benefits from tailor-curated feedback data. These ablation results underscore the crucial role played by our proposed identity consistency and aesthetic rewards in achieving high-quality, identity-preserving image generation.

\begin{figure}[!t]
    \centering
    \includegraphics[width=1.0\linewidth]{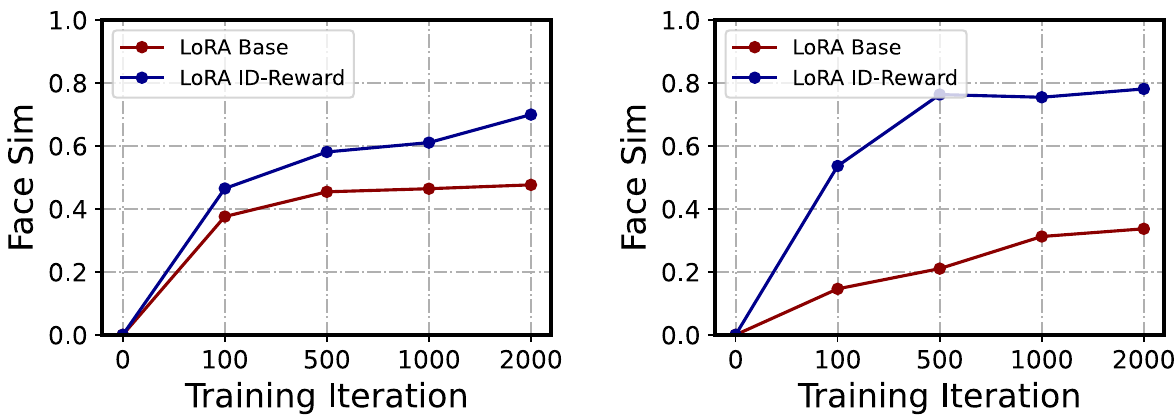}
     \vspace{-10pt}
    \caption{The illustration of the accelerated identity adaptation for the LoRA model. \textbf{Left}: LoRA trained based on SD1.5. \textbf{Right}: LoRA trained based on SDXL.}
    \label{fig:fast_adaptation}
\end{figure}

\noindent\textbf{Fast Identity Adaptation:} The LoRA approach is a test-time fine-tuning method and needs to train a separate LoRA model for each portrait. This poses a significant challenge for the application as it requires enough training time to ensure adequate identity adaptation.  Thanks to the targeted feedback on identity consistency, we found our method can accelerate the identity adaptation of LoRA training significantly as demonstrated in Fig.\ref{fig:fast_adaptation}. This effect is particularly prominent when adapting to the SDXL, as conventional LoRA adaptation for SDXL is inherently slow due to the larger number of parameters required to update. In contrast, the id-aligner considerably reduces the fine-tuning time to achieve the same level of face similarity.

\noindent\textbf{Generalization Study:} To demonstrate the generalization ability of our approach, We utilized the widely recognized Dreamshaper\footnote{https://huggingface.co/Lykon/DreamShaper} and RealVisXL\footnote{https://huggingface.co/SG161222/RealVisXL\_V3.0} for the open-sourced alternatives of SD15 and SDXL, and validate our method with these text-to-image models. According to the results of Tab.\ref{tab:generalization_study}, our method delivers a consistent performance boost on these alternative base models. Specifically, our method brings \textbf{10.7\%} and \textbf{4.2\%} performance boosts with Dreamshaper in terms of face similarity and image similarity measured by CLIP, which means better identity preservation. Moreover, our method obtained more significant performance improvement with the more powerful model in SDXL architecture. For instance, our method surpasses the original RealVisXL model with \textbf{22.3\%} in identity preservation, and \textbf{8.3\%} improvements in CLIP-I. This demonstrates the superior generalization ability of our method on different text-to-image models.

\begin{figure}[!t]
        \centering
        \includegraphics[width=1\linewidth]{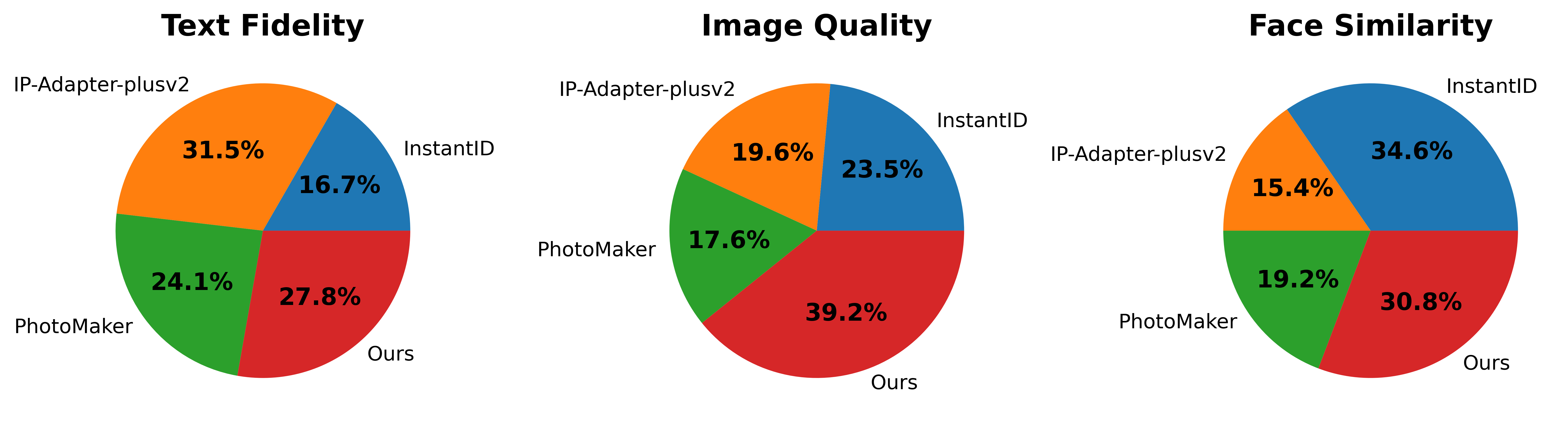}
        \caption{User preferences on Text fidelity,  Image quality, Face Similarity for different methods. We visualize the proportion of total votes that each method has received. }
        \label{fig:user_study}
\end{figure}

\subsubsection{\textbf{User Study}}
To gain a comprehensive understanding, we conducted a user study to compare our method with IP-adapter-plusv2 \cite{ipadapter}, PhotoMaker \cite{photomaker}, and InstantID \cite{instantid}. We presented 50 generated text-image pairs and a reference face image to each user. For each set, users were asked to vote for the best one or two choices among the four methods, based on three criteria: (1) text-fidelity - which image best matches the given prompt, (2) Image Quality - which image looks the most visually appealing, and (3) Face similarity - which image most closely resembles the reference face. Users could choose two options if it was difficult to select a clear winner. We collected a total of 500 votes from 10 users.

As shown in Fig. \ref{fig:user_study}, the results align with the quantitative study in Fig. \ref{fig:comparison}. InstantID achieved the highest face similarity score, while our method secured the second-best face similarity score. Our method obtained the highest aesthetic score and the second-highest text-image consistency score. Overall, our method performed well across all indicators and exhibited a relatively balanced performance compared to other methods.



%% file: sec/5_conclusion.tex
\section{Conclusion}
In this paper, we introduced \textbf{ID-Aligner}, an algorithm crafted to optimize image generation models for identity fidelity and aesthetics through reward feedback learning. We introduces two key rewards: identity consistency reward and identity aesthetic reward, which can be seamlessly integrated with adapter-based and LoRA-based text-to-image models, consistently improving identity consistency and producing aesthetically pleasing results. Experimental results validate the effectiveness of ID-Aligner, demonstrating its superior performance. 

\clearpage
